%% file: paper.tex
\documentclass[runningheads]{llncs}
\usepackage[T1]{fontenc}
\usepackage{graphicx}

\usepackage{amsmath,amsfonts}
\usepackage{graphicx}
\DeclareMathOperator*{\argmax}{arg\,max}

\usepackage{graphicx}
\usepackage{hyperref}
\usepackage{url}
\usepackage{float}
\usepackage{setspace}
\usepackage{harpoon}
\usepackage{multicol}
\usepackage{multirow}
\usepackage{caption}
\usepackage{hhline}
\usepackage{subcaption}
\usepackage{enumitem}
\usepackage{url}
\usepackage{lipsum}  

\newcommand{\etal}{\textit{et al.}}

\begin{document}
\title{Adaptive Attractors: A Defense Strategy against ML Adversarial Collusion Attacks}
\titlerunning{A Defense Strategy against ML Adversarial Collusion Attacks}

\author{Jiyi Zhang \and Han Fang \and Ee-Chien Chang}
\authorrunning{J. Zhang et al.} %
\institute{School of Computing, National University of Singapore\\
\email{{jiyizhang, fanghan}@u.nus.edu},\\\email{changec@comp.nus.edu.sg}}

\maketitle              %
\input{abstract.tex}

\input{intro}
\input{background}
\input{threatmodel}
\input{approach}

\input{analysis}
\input{eval}

\input{conclusion}

\clearpage
\bibliographystyle{splncs04}
\bibliography{references}

\end{document}

%% file: abstract.tex
\begin{abstract}
    In the seller-buyer setting on machine learning models, the seller generates different copies based on the original model and distributes them to different buyers, such that adversarial samples generated on one buyer’s copy would likely not work on other copies. A known approach achieves this using attractor-based rewriter which injects different attractors to different copies. This induces different adversarial regions in different copies, making adversarial samples generated on one copy not replicable on others. In this paper, we focus on a scenario where multiple malicious buyers collude to attack.  We first give two formulations and conduct empirical studies to analyze effectiveness of collusion attack under different assumptions on the attacker’s capabilities and properties of the attractors. We observe that existing attractor-based methods do not effectively mislead the colluders in the sense that adversarial samples found are influenced more by the original model instead of the attractors as number of colluders increases (Figure~\ref{fig:multicopies}). Based on this observation, we propose using adaptive attractors whose weight is guided by a U-shape curve to cover the shortfalls. 
    Experimentation results show that when using our approach, the attack success rate of a collusion attack converges to around 15\% even when lots of copies are applied for collusion. In contrast, when using the existing attractor-based rewriter with fixed weight, the attack success rate increases linearly with the number of copies used for collusion.
\end{abstract}

%% file: intro.tex
\section{Introduction}
Machine learning models have become ubiquitous in many applications, including image and speech recognition, natural language processing, and autonomous driving. However, these models are vulnerable to adversarial attacks~\cite{DBLP:conf/kdd/LowdM05,DBLP:conf/iclr/BrendelRB18,Kurakin2017AdversarialEI}, where an attacker deliberately modifies the input data to cause misclassification or other unintended behaviors. 

While direct defense~\cite{Shaham2018UnderstandingAT,Parseval_Networks,DBLP:journals/corr/abs-1802-00420} of adversarial attack is extremely difficult, one solution to mitigate these attacks is to deploy different copies of the model to different users, so that an adversarial sample generated on one copy will not replicate on another copy~\cite{DBLP:journals/corr/abs-2111-15160}. One efficient method of creating different copies of a model without re-training is through parameter rewriting. One example of the rewriter is based on attractors~\cite{DBLP:journals/corr/abs-2003-02732}. Injecting different attractors can induce different adversarial regions in different copies with low overlaps. Adversarial samples drawn from non-overlapping adversarial regions are not replicable.

However, the above solution faces a problem in the scenario where the adversaries collude instead of attacking the model independently. In existing approaches, the success rate of a collusion attack increases linearly with the number of colluders.

In this paper, we introduce a strategy to mitigate collusion attacks in multi-copy defenses. Our approach builds upon the concept of parameter rewriting. However, instead of utilizing a static weight for the rewriter, we employ a dynamic weight that varies based on the proximity of an input sample to the decision boundary. To determine the weight value, we employ a U-shaped curve guided by several insights derived from our observations during simulations and experiments. By adopting this flexible weight mechanism, we aim to enhance the robustness of our defense approach against collusion attacks.

The empricial analysis conducted demonstrates the resilience of our proposed method in scenarios involving many copies used for collusion. Through actual attack experiments, we observed a substantial reduction in the success rate of both single-copy replication attacks and multi-copy collusion attacks when employing our method. For example, when one of the collusion attacks~\cite{DBLP:journals/corr/abs-2111-15160} is applied to our approach, the attack success rate converges to approximately 15\%, even when a large number of copies are employed. In contrast, when the same attack is applied to the original attractor-based rewriter, the attack success rate exhibits a linear increase as the number of colluding copies rises.
\\ \\
\noindent
{\em \bf Contributions:}
\begin{enumerate}[leftmargin=*]
    \item We propose an approach based on parameter rewriting and attractors to mitigate collusion attacks.
    \item We introduce a flexible weight mechanism for the rewriter component based on a sample's distance to the decision boundary.
    \item We utilize a U-shaped curve to guide the weight assignment based on several intuitions derived from observations made during simulations.
    \item Experimentation results show a significant reduction in the success rate of both single-copy replication attacks and multi-copy collusion attacks when employing the proposed method.
\end{enumerate}

%% file: background.tex
\section{Background and Related Works}
\noindent
{\em Adversarial attacks.}\ \
Adversarial attacks are deliberate techniques used to deceive and manipulate machine learning models~\cite{Szegedy2014IntriguingPO}. By introducing carefully crafted input samples, adversarial attacks exploit vulnerabilities in the model's decision-making process, leading to incorrect predictions or misclassifications. These attacks often involve adding subtle and imperceptible perturbations to the input data, aiming to bypass the model's defenses and compromise its integrity.
Adversarial attacks typically generate samples by exploring directions based on local properties~\cite{Szegedy2014IntriguingPO}. For instance, the Fast Gradient Sign Method (FGSM) method~\cite{Goodfellow2015ExplainingAH} derives the search direction from the gradient of the training loss function.
\\ \\
\noindent
{\em Attractors.}\ \
Attractor~\cite{DBLP:journals/corr/abs-2003-02732} is a proactive approach which detects adversarial perturbations by considering characteristics of the attack process, which is different from traditional methods that consider the characteristics of the adversarial samples~\cite{DBLP:journals/corr/GrosseMP0M17,DBLP:journals/corr/GongWK17,DBLP:journals/corr/abs-1806-00081}.
The key idea is to confuse and mislead the attacker.
This method treats each attack as an optimization problem, where the goal of the attractor is to actively modify the optimization objectives within the classifier model. 
By injecting attractors into the model, it effectively introduces artifacts that taint the local properties of the search space, causing the gradients of the attack objective function to point towards a specific attractor in the neighborhood, thereby confusing the adversary's search process and luring it towards dedicated regions in the space. As a result, adversarial samples are generated from these dedicated regions and become detectable.

The concept of confusing the adversary has been extended in other works~\cite{DBLP:conf/nips/ChenHTWXH22,DBLP:journals/corr/abs-2106-04938}.
For example, Chen~\etal\cite{DBLP:conf/nips/ChenHTWXH22} focused on score-based black-box settings and proposed a post-processing method which modifies logits and misleads the adversary with loss attractors. Wu~\etal\cite{DBLP:journals/corr/abs-2106-04938} proposed Hedge Defense which adds an extra layer of pre-processing and alters the input sample with specifically designed noise. This mechanism is able to perturb an adversarial sample and neutralizes the adversary's attempt to modify classification predictions.
\\ \\
\noindent
{\em Parameter Rewriting and Distribute Different copies to Different Users.}\ \
Inspired by the effectiveness of attractors, a recent study~\cite{DBLP:journals/corr/abs-2111-15160} proposed the utilization of attractors as a parameter rewriting mechanism to generate diverse copies of an original model, which can be distributed to different buyers.

The attractor-based rewriter operates by modifying the predictions of the original model, introducing distinctive steep bumps and holes onto the initially smooth surface. 
In each individual copy, the positions and sizes of the holes and bumps differ. These variations are controlled by unique hyper-parameters assigned to each specific copy. As a result, the adversarial search process leads to distinct adversarial regions across different copies. Therefore, adversarial samples discovered in one copy are unlikely to be effective on another copy.

The integration of attractors as a rewriting mechanism introduces diversity in the distributed copies of the original model. By creating unique adversarial regions in each copy, the approach effectively thwarts attacks that rely on adversarial samples being replicable between different copies of the model.

The concept of distributing different copies to different buyers leads to many other applications. For example, Fang \etal\cite{DBLP:journals/corr/abs-2301-01218} also considered the buyers-seller setting and proposed a tracing framework which injects unique properties into each buyer's copy such that adversarial examples generated on different copies have different characteristics and can be traced to their source copies.

%% file: threatmodel.tex
\section{Threat Model}
We adopt the same buy-seller distribution setting as Zhang~\etal~\cite{DBLP:journals/corr/abs-2111-15160}.
We denote the original copy as ${\mathcal M}_\beta$. For each buyer $i$, the seller generates and distributes a unique copy ${\mathcal M}_{\beta_i}$. The generation process is through parameter rewriting which is defined in Zhang~\etal~\cite{DBLP:journals/corr/abs-2111-15160}, and does not involve re-training. Each buyer has black-box and hard-label access to its own copy.

\paragraph{Adversary Capabilities.}\ \
The adversary has the ability to gain black-box and hard-label access to one or multiple copies of the model. The adversary can interact with the models independently or simultaneously.

\paragraph{Attack Objective.}\ \
The adversary wants to compromise a victim buyer ${\tt vic}$ 's model ${\mathcal M}_{\beta_{\tt vic}}$.  Given a sample $\textbf x$, the adversary's goal is to generate a new sample $\textbf x'$ such that classes predicted by ${\mathcal M}_{\beta_{\tt vic}}$ are different for $\textbf x'$ and $\textbf x$. Specifically:
\begin{equation}
\label{eq:argmax}
\argmax {\mathcal M}_{\beta_{\tt vic}} (\textbf x) \not =\argmax {\mathcal M}_{\beta_{\tt vic}} (\textbf x').     
\end{equation}
In addition, the difference between $\textbf x'$ and $\textbf x$ should be as small as possible.

\paragraph{Replication Attack.}\ \
The adversary carries out the attack against its own copy then apply the adversarial samples on the victim's copy in hopes that the attack can replicate.

\paragraph{Collusion Attack.}\ \
The adversary aims to launch an attack by collating information from different copies of the model. Through collusion, the adversary seeks to diminish the impact of parameter rewriting techniques, and achieve maximum attack success rate.

%% file: approach.tex
\section{Analysis of Attractors}
This section gives two formulations to estimate effectiveness of collusion attacks. Beside providing an estimate  for large $n$, the number of colluders, these formulations also provide insights for further improvement.

\subsection{Formulation of Copies Obtained through Parameter Rewriting}
In this section, we begin by formulating the parameter rewriting method utilizing probabilistic models. Subsequently, we conduct simulations to estimate the bounds for collusion attacks and compare them with the actual success rate of collusion attacks against the original attractor-based defense. Through this analysis, we gain insights into the characteristics of replication attacks and identify crucial areas for improvement.

We give two formulations. The first one is based on `summation' construction and sums two random variables together. The second one is directly based on the mechanism of attractors, that is, different adversarial regions are induced in different copies.

\paragraph{Formulation 1: Copies Obtained through Summation with Rewriters.}\ \
In Zhang~\etal's approach~\cite{DBLP:journals/corr/abs-2111-15160}, the rewriting component is combined with the original model through direct summation. 

To formulate this construction, we first define a random variable $O$ with standard normal distribution. Let $O$ be associated with the original copy. 

Then for each buyer $i$, we define a random variable $R_i$ with standard normal distribution. We then define $C_i$ as a weighted sum of $O$ and $R_i$, that is $\eta O + (1-\eta) R_i$, where $\eta$ is a pre-defined constant. Let $C_i$ be associated with the copy of buyer $i$.

In this formulation, the random variable is interpreted as a function which maps a copy to the `effectiveness' of a given attack against this copy. Randomness is commonly introduced in the attack process through techniques such as random noise generation, random perturbation selection, or random initialization of optimization algorithms. By incorporating randomness, adversaries aim to explore different directions or search spaces to find adversarial samples that can evade detection or mislead classification. For a buyer $i$, we define the attacks to be successful on copy $i$ if the attack effectiveness exceeds a pre-defined threshold $t$. In other words, the probability that $C_i$ lies in the interval $(t, \infty)$ is the success rate of the attack.

\paragraph{Replication Attack.}\ \
Suppose the attacker conducts attack on his own copy $C_{atk}$ first then applies the successful samples to a victim copy $C_{vic}$, we define the success rate of replication attack as:

$$Pr(C_{vic} > t\ |\ C_{atk} > t)$$

Suppose the attacker generates a sample that can attack $n$ copies that the attacker owns, then the success rate of replication attack against the victim copy is:

$$Pr(C_{vic} > t\ |\ C_{atk_1} > t, C_{atk_2} > t,...,C_{atk_n} > t)$$

\paragraph{Simulation Construction.}\ \
To simulate the scenario, we employ the Monte Carlo method. Initially, an array is generated using the random variable $O$, with its length corresponding to the number of attack samples. For each copy $i$, a distinct array is generated using the random variable $R_i$. Subsequently, the summation operation outlined in the aforementioned formulation is applied to create an array of the same length, representing the combined copy associated with $C_i$.

Suppose $\mathbf{vic}$ is the array representing the victim and $\mathbf{atk}$ is the array representing the attacker. Each of them contains $n$ elements. If $\mathbf{atk}[k] > t$ and $\mathbf{vic}[k] >t$, means the replication attack using sample $k$ is successful. Then 
$$\frac{\text{Number of } k \text{ such that } \mathbf{vic}[k]>t \text{ and } \mathbf{atk}[k]>t}{\text{Number of } k \text{ such that } \mathbf{atk}[k]>t}$$
is the overall success rate for replication attacking using $n$ samples.

For collusion attack, the $\mathbf{atk}$ becomes a 2D array with number of available copies $m$ as the additional dimension. The overall success rate then becomes:
$$\frac{\text{Number of } k \text{ such that } \mathbf{vic}[k]>t \text{ and } \mathbf{atk}[l][k]>t, \text{ for } l \in [0,m)}{\text{Number of } k \text{ such that } \mathbf{atk}[l][k]>t, \text{ for } l \in [0,m)}$$

\paragraph{Formulation 2: Copies Obtained through Embedded Attractors.}\ \
For attractor-based rewriters, instead of using random variables to formulate the problem, we construct an alternative formulation directly based on the motivation that each copy has different adversarial regions.

We use $O$ to denote the original model. It is in the form of a set. Each entry in the set is a tuple which contains a point in $d$-dimensional space and a radius value. For example, if the space is 3 dimensional, the tuple will be in the form of $((x,y,z), r)$. It represents a $3$-dimensional sphere with radius $r$ centered at point $(x,y,z)$. This sphere represents one adversarial region in the original model. Therefore, the whole set $O$ contains representations of all adversarial regions in the original model. The values of $x,y,z,r$ in each entry are independently chosen from uniform distributions.

A copy issued to buyer $i$ is embedded with attractor $A_i$. $A_i$ is in the same form as $O$ but represents new adversarial regions induced by the injected attractors. We denote this copy as $C_i$ where $C_i = O \cup A_i$.

\paragraph{Simulation Construction.}\ \
We also utilize the Monte Carlo method for simulation. For each attempt of replication attack, we randomly sample a point in the adversarial regions of attacker's copy $C_{atk}$, if the same point is also contained in the adversarial region of victim copy $C_{vic}$, then the attack is successful. 

If the point is in adversarial regions of the original model $O$, then the attack should be successful on all copies. If the point is in adversarial regions of $A_{atk}$, it can attack $C_{vic}$ if the attractor-induced adversarial regions have overlaps. During a collusion attack, the attacker samples a point in the overlapping region of adversarial regions across all available colluding copies.

\subsection{Simulation}
We conduct simulation for both formulations under collusion attack.
We also compare the simulation performance with the actual performance of attractor-based parameter rewriting under collusion attack. For this experiment on actual attack, the construction of attractors is exactly same as Zhang~\etal's attractor-based rewriter using QIM decoder~\cite{DBLP:journals/corr/abs-2111-15160}. We also employ the same custom collusion attack which is modified based on DeepFool~\cite{MoosaviDezfooli2016DeepFoolAS}.

For parameters of simulations, we choose them in such a way that the replication attack success rate will have the same starting point (where the adversary only has access to one copy) as the actual experiment results.

The results are present in Figure~\ref{fig:simulation}. The horizontal axis represents the number of copies used in the collusion attack. The vertical axis represents the attack success rate. From the results, we can observe that the actual performance of attractor-based rewriter can be approximated using formulation 2. This result shows that the performance of attractor-based rewriter using QIM decoder is in line with the theoretical performance of attractor-based parameter rewriting through injecting holes and bumps. 

On the other hand, formulation 1 demonstrates a much higher attack success rate when the number of copies are small. This gap between the simulation result and actual performance may indicate that the actual adversarial attacks are only able to find a small portion of all the adversarial samples in the space. In contrast to random sampling, objective function-guided adversarial attacks exhibit higher efficiency in finding adversarial samples. However, they often come at the cost of reduced sample quantity.
While adversarial samples theoretically exist throughout the input space, uniformly sampling to find them proves challenging for attackers. In our simulation, we observed that with formulation 1, it becomes increasingly difficult to discover an adversarial sample that is effective across more than 17 attack copies.

Although the collusion attack cannot be extended to a larger number of copies, we can observe that formulation 1 exhibits convergence below 20\%. This presents an opportunity to develop a parameter rewriting method that surpasses the performance of the original attractor-based rewriter and achieves a success rate increase that is less than linear in relation to the number of attack copies.
\input{figures/simulation.tex}
\subsection{Analysis of Shift in Each Component of a Copy under Attack}
\label{sec:shift}
The observation that the actual performance of attractor-based rewriter fits well with simulation under formulation 3 suggests that, when under attack, the change of score in victim class is either contributed mostly by the original model or mostly by the attractor component, and they seldom have equal contribution. 

To verify this hypothesis, we conduct an experiment to analyze the shift of score in the original model component and the rewriter component when under attack. In this experiment, we conduct adversarial attacks on 10,000 samples from CIFAR-10 dataset~\cite{krizhevsky2009learning}. The model uses the attractor-based rewriter with QIM (Quantization Index Modulation) decoder.
The adversarial attack we choose is a custom collusion attack from Zhang~\etal's work\cite{DBLP:journals/corr/abs-2111-15160}.
For each successful attack, we plot the change of score in victim class. The change is presented separately for the original model and rewriter component. We repeat the experiment for two scenarios: replication attack using a single copy and collusion attack using eight copies.
The results are shown in Figure~\ref{fig:multicopies}.
\input{figures/multicopies.tex}
The horizontal axis represents the change contributed by the attractor-based rewriter component. The vertical axis represents the change contributed by the original model. In Figure~\ref{fig:multicopies}, we can observe that the evidence supports the hypothesis, that is, the points form two clusters with few points in between. This shows that the score change caused by the attack most likely concentrates on one component, instead of contributed by both original model and attractor.

The larger cluster on the horizontal axis suggests that the attractor component indeed attracts the attack. However, we observe a smaller cluster near the origin. This suggests that the attack is likely to cause more impact on the original model when the input sample is near the decision boundary. 

When using eight copies for the collusion attack, we observe that the center of the large cluster shifts toward left. This shows that the influence of attractor is partially mitigated by collusion. At eight copies, the magnitude of shift is not large but still obvious. This observation is in line with the linear increase of attack success rate observed in Figure~\ref{fig:simulation}.

\subsection{Area of Improvement}
\label{sec:improve}
The observation in Section~\ref{sec:shift} suggests that adding attractors in a more strategic way would be a potential measure to significantly mitigate collusion attack. We give the details of our proposed improvement in Section~\ref{sec:approach}. Here, we explain the intuitions that lead to our new approach.
\\ \\
\noindent
{\em Intuition 1: More attractor influence near the decision boundary.}\ \
Figure~\ref{fig:multicopies} suggests that the attacker is more likely to cross the decision boundary of the original model when the input sample is near the decision boundary. One potential measure to mitigate this is to enhance the influence of attractors near the decision boundary. Ideally, most clean samples are far away from the decision boundary, so the overall accuracy should not be affected too much. 
Given a higher attractor weightage, the attacker will be more likely affected, and it will be more difficult to avoid the influence of attractors.
\\ \\
\noindent
{\em Intuition 2: More attractor influence in regions far away from the decision boundary.}\ \
For a clean sample that is far away from the decision boundary, the difference in scores of the highest and second-highest class is usually very large ($\geq$ 0.9). This allows more room of attractor influence. For example, the weight of the attractor can be set as $\mu \times(1 - \epsilon)$ where $\mu$ is a constant that minimizes the gap between the highest score and second-highest in the original classifier to 0 and $\epsilon$ is a very small value. For simplicity, we set $\mu$ as:
\begin{equation}
\label{eq:mu}
\mu = \frac{1st\ highest\ - 2nd\ highest\ in\ {\mathcal M}_{\beta_{i}}({\textbf x})}{max({\mathcal A}_{k_{i}}({\textbf x}))-min({\mathcal A}_{k_{i}}({\textbf x}))}
\end{equation}
\noindent
{\em Intuition 3: Exploit the iterative nature of attacks.}\ \
Most adversarial attacks are carried in an iterative manner, so the adversarial sample can change prediction result while staying near to the original clean input sample. The existing approach places attractors in fixed positions in the input space and the objective function of an attack guides it to move closer to a nearby attractor. If we can adaptively change the weight of the rewriter according to input sample's distance to decision boundary, the objective function of the attack will be following a moving target in each iteration. Instead of ending up in a local minimum, the attack may move in circles and unable to converge within a reasonable number of iterations.

\section{The Proposed Approach: Adaptive Attractors}
\label{sec:approach}
Ideally, the effectiveness of the parameter rewriting should be maintained at a constant level which is independent of the number of copies available to the adversary. If a method can meet this requirement, it is likely much less vulnerable to collusion attack. In this section, we discuss the improvement we propose to defend against collusion attack.

Drawing inspiration from the original concept of attractors, our construction also employs the "summation" method. In our approach, the generated model ${\mathcal M}_{\beta_i}$ is also constructed by combining the original model copy with the attractor-based rewriter.

More specifically, when presented with an input ${\textbf x}$, the output of ${\mathcal M}_{\beta_i}$ is obtained through the normalized sum (with respect to the L1 norm) of the outputs from both components, that is:

\begin{equation}
    \label{eq:nsum}
    {\mathcal M}_{\beta_{i}}({\textbf x}) = \frac{ {\mathcal M_{\beta}} ({\textbf x}) + \alpha {\mathcal A}_ {k_{i}}({\textbf x}) } { \| {\mathcal M_{\beta}} ({\textbf x}) + \alpha {\mathcal A}_{k_{i}}({\textbf x})   \|_1}
\end{equation}
where $\alpha = {\mathcal F}({\bf x}, {\mathcal M}_{\beta}, {\mathcal A}_{k_{i}})$

Instead of using a fixed constant to determine the weightage of attractor injected to each copy, we set $\alpha$ as a flexible value, which is computed using the function $\mathcal F$ and three factors: input ${\textbf x}$, the original model ${\mathcal M_{\beta}}$ and the attractor-based rewriter ${\mathcal A}_{k_{i}}$.

\subsection{U-shape curve for a flexible $\alpha$ value}
\label{sec:u}
Based on the intuitions in Section~\ref{sec:improve}, we propose to employ a flexible $\alpha$ value which follows a U-shape as shown in Figure~\ref{fig:ushape}. More attractors are added to regions near and far away from the decision boundary. 

In Figure~\ref{fig:ushape}, we define the distance to decision boundary as the different between the scores of the highest class and second-highest class. Note that this value is computed after applying the softmax function to the output of the model. Therefore, the range is from 0 to 1.

When determining the actual value of $\alpha$, we have two major considerations: values at two ends of U-shape and values in the middle of the U-shape. 

For a sample $\bf x$ which is far away from the decision boundary (the right end of the U-shape curve), we first derive a value for the weight of rewriter such that the predicted class for the combined model ${\mathcal M}_{\beta_{i}}$ is about to flip from the prediction made by the original model ${\mathcal M_{\beta}}$. We denote this value as $\mu$. It can be computed by solving Equation~\ref{eq:argmax}. Here, we use the simplified version as stated in Equation~\ref{eq:mu} without considering attractor's correlation to each class. We then compute $\alpha = \mu \times(1 - \epsilon)$ where $\epsilon$ is a very small value. By doing this, the prediction of the combined model ${\mathcal M}_{\beta_{i}}$ stays the same as the original model ${\mathcal M_{\beta}}$.

For a sample $\bf$ which is near the decision boundary (the left end of the U-shape curve), we choose the weightage of attractors empirically. In our example, we work on CIFAR-10 dataset. We feed the 50,000 training images of CIFAR-10 into the model and select all the images with their distance to decision boundary less than 0.2. We allocate them into two intervals according to their distance to decision boundary: from 0 to 0.1 and from 0.1 to 0.2. For each interval, we choose the maximum $\alpha$ value such that the overall average trade off in accuracy is 0.5\% (for this selected set of images).

For a sample from the middle ground (at the bottom of the U-shape curve), we use a pre-defined $\alpha$ value that is slightly above 0.
\\ \\
\noindent
{\em Remarks.}\ \
To see how flexible weight affects an attack, we consider from the perspective of the adversary. On one hand, when the starting point of an attack is near or very far away from the decision boundary. The attack will be misguided by the enhanced influence of attractors and end up in local minima. On the other hand, if the adversary uses some other objective functions which intentionally avoid attractors, it will likely end up circling around regions in the middle ground (the bottom of the U-shape), therefore unable to cross over to the other side of the decision boundary, making the attack unsuccessful.
\input{figures/ushape.tex}

%% file: figures/simulation.tex
\begin{figure}[H]
  \centering
  \vspace{-15pt}
      \centering
      \includegraphics[width=0.6\linewidth]{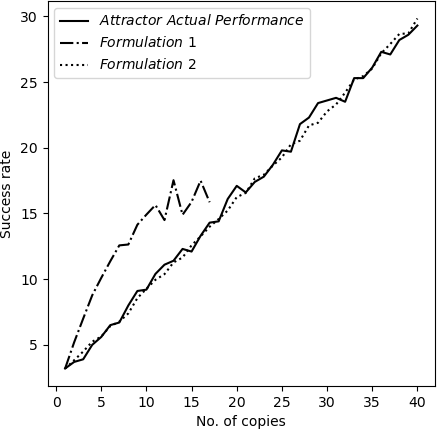}
    \caption{Compare simulation results with actual performance of attractor-based parameter rewriting.}
    \label{fig:simulation}
  \end{figure}
  \vspace{-15pt}

%% file: figures/multicopies.tex
\begin{figure}[H]
  \vspace{-15pt}
    \centering
    \begin{subfigure}{.5\linewidth}
      \centering
      \includegraphics[width=0.95\linewidth]{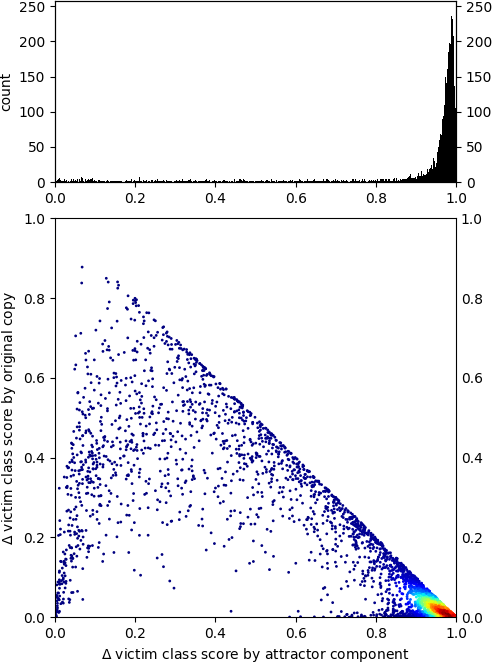}
      \caption{1 copy.}
    \end{subfigure}%
    \begin{subfigure}{.5\linewidth}
      \centering
      \includegraphics[width=0.95\linewidth]{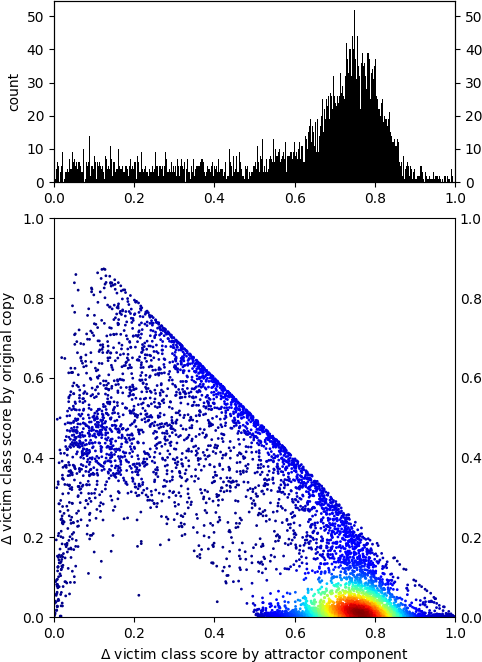}
      \caption{8 copies.}
    \end{subfigure}
    \caption{Shift caused by increasing number of colluding copies.}
    \label{fig:multicopies}
\end{figure}
\vspace{-15pt}

%% file: figures/ushape.tex
\begin{figure}[H]
  \vspace{-15pt}
  \centering
      \centering
      \includegraphics[width=0.7\linewidth]{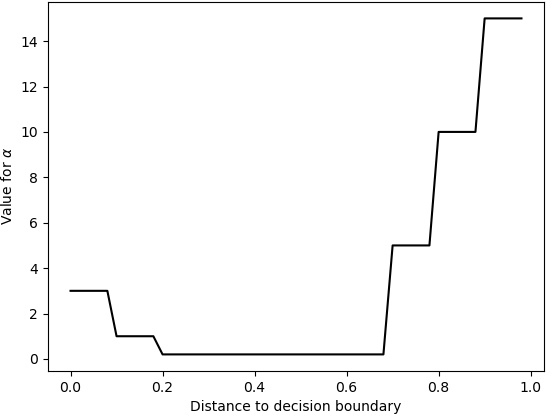}
    \caption{Illustration for flexible $\alpha$ following a U-shape curve.}
    \label{fig:ushape}
  \end{figure}
  \vspace{-15pt}

%% file: analysis.tex
\subsection{Analysis of Shift in Each Component of a Copy under Attack}
We conduct the same experiment as Figure~\ref{fig:multicopies} while using the flexible $\alpha$ we propose in Section~\ref{sec:u}. The results are shown in Figure~\ref{fig:improve2}, 

From Figure~\ref{fig:improve2}, the first observation is that the attack is less likely to succeed when flexible $\alpha$ is applied. There are fewer points in the figure, indicating that there are fewer samples becoming adversarial after the attack.

For successful samples, we can observe that the shift in scores also form two clusters. For both clusters, the dominant part of the score change is contributed by the attractor-based rewriter. In addition, the change of score in the rewriter component form two peaks around 0.5 and 1.0 respectively. They correspond to the two ends in the U-shape curve which we define in Section~\ref{sec:u}.

Another observation is that shift no longer varies much with the increasing number of copies in collusion attack. The positions of the peaks on horizontal axis remain the same when the copies used for attack increase from one to eight. This suggests that the flexible $\alpha$ approach could be more resistant against collusion attacks.

\input{figures/improve2.tex}

%% file: figures/improve2.tex
\begin{figure}[H]
  \vspace{-15pt}
    \centering
    \begin{subfigure}{.5\linewidth}
      \centering
      \includegraphics[width=0.95\linewidth]{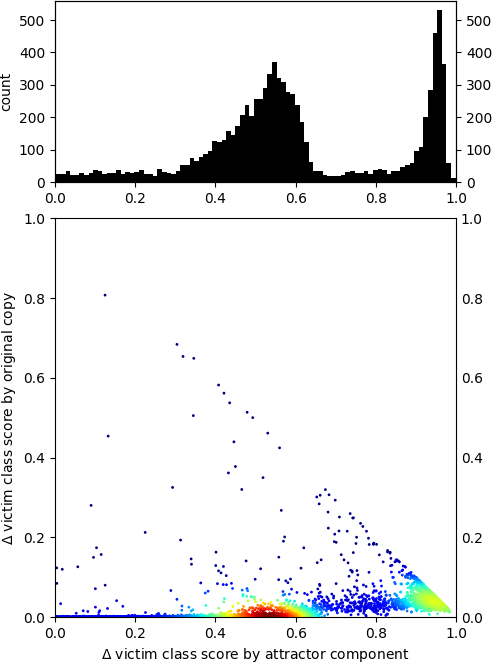}
      \caption{1 copy.}
    \end{subfigure}%
    \begin{subfigure}{.5\linewidth}
      \centering
      \includegraphics[width=0.95\linewidth]{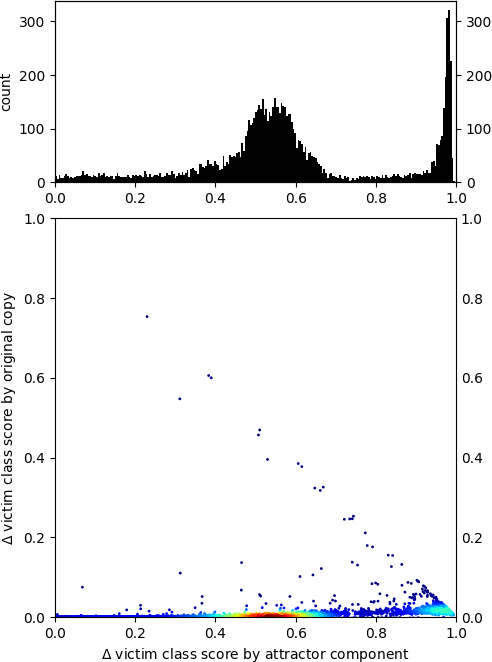}
      \caption{8 copies.}
    \end{subfigure}
    \caption{Shift caused by increasing number of colluding copies (with flexible $\alpha$).}
    \label{fig:improve2}
\end{figure}
\vspace{-15pt}

%% file: eval.tex
\section{Evaluation}

\subsection{Setup}
\noindent
{\em Datasets.}\ \
We conducted our evaluation on two distinct datasets, namely CIFAR-10 (Canadian Institute For Advanced Research)\cite{krizhevsky2009learning} and GTSRB (German Traffic Sign Recognition Dataset)\cite{Houben-IJCNN-2013}. CIFAR-10 consists of a diverse range of images, including animals and vehicles, with a training set comprising 50,000 images and a testing set of 10,000 images. On the other hand, GTSRB focuses on traffic signs and includes 43 different classes. Its training set comprises 39,209 images, while the testing set contains 12,630 images. To ensure consistency, all images were uniformly resized to dimensions of $32 \times 32 \times 3$.
\\ \\
\noindent
{\em Models.}\ \
In order to have a fair comparison with the original attractor-based rewriter, we apply the same setting for the models as Zhang~\etal~\cite{DBLP:journals/corr/abs-2111-15160}.
\\ \\
We first train two original models with the same network architecture from scratch and use them as references. In Zhang~\etal's work~\cite{DBLP:journals/corr/abs-2111-15160}, this is named as independent-training method.
\begin{itemize}[topsep=5pt]
    \item $\mathcal{M}_{\phi}$: Classifier trained randomly from scratch and parametrized by $\phi$.
    \item $\mathcal{M}_{\psi}$: Classifier trained randomly from scratch and parametrized by $\psi$.
\end{itemize}
\noindent
For both the original approach and the new approach with flexible weight, we construct three other models on top of the reference models for the experiments:
\begin{itemize}[topsep=5pt]
    \item $\mathcal{M}_{\phi_1}$: $\mathcal{M}_{\phi}$ using attractor-based rewriter ${\mathcal A}_{k_1}$.
    \item $\mathcal{M}_{\phi_2}$: $\mathcal{M}_{\phi}$ using attractor-based rewriter ${\mathcal A}_{k_2}$.
    \item $\mathcal{M}_{\psi_2}$: $\mathcal{M}_{\psi}$ using attractor-based rewriter ${\mathcal A}_{k_2}$.
\end{itemize}

For the architecture of the models, we apply the same VGG-19~\cite{DBLP:journals/corr/SimonyanZ14a} as Zhang~\etal~\cite{DBLP:journals/corr/abs-2111-15160}. The training of reference model also takes 200 epochs. For attractor-based rewriters, QIM decoder is used for both the original method and the new proposed method. The only difference is the fixed and flexible weight of the rewriter.

\subsection{Accuracy of the Rewritten Copies}
We evaluate the accuracy of the copies rewritten by the original attractor-based rewriter and the new adaptive rewriter. The evaluation is conducted on the testing datasets of CIFAR-10 and GTSRB. The results are shown in Table \ref{tab:onclean}. 

\begin{table}[H]
    \vspace{-10pt}
    \centering
    \small
    \begin{tabular}{|c|c|c|} 
        \hline
        Datasets                                                                            & CIFAR-10          & GTSRB              \\ 
        \hline
        Original Copy                                                                       & 96.3\%  & 95.8\%   \\ 
        \hline
        \begin{tabular}[c]{@{}c@{}}Rewritten with Attractor (QIM)\\(Fixed Weight)\end{tabular}             & 94.5\%  & 95.3\%   \\ 
        \hline
        \begin{tabular}[c]{@{}c@{}}Rewritten with Attractor (QIM) \\(Flexible Weight)\end{tabular}                                                          & 95.1\%   & 95.4\%   \\
        \hline
        \end{tabular}
    \caption{Accuracy of original model and two new copies rewritten by the original attractor-based rewriter and new adaptive rewriter respectively. }
    \label{tab:onclean}
\end{table}
\vspace{-20pt}

The results show that the accuracy trade off in the new adaptive rewriter approach is smaller than that of the original attractor-based rewriter. This is in line with the design of the adaptive attractor. By using the U-shape curve to determine the weight of rewriter, the overall amount of attractors added to the model is actually reduced. Though more weightage is given to attractors in regions near and far from the decision boundary, they either do not change the prediction result by design (for far away regions) or has their weight bounded (for regions near boundary), thus have minimal impact to the model's overall accuracy.

\subsection{Performance under Attack}
In this section, we conduct the evaluation of standard replication attack (without collusion) and compare the result with the original attractor-based rewriter. Here, we use boundary attack~\cite{DBLP:conf/iclr/BrendelRB18}, Hop Skip Jump attack (HSJ)~\cite{DBLP:conf/sp/ChenJW20} and Geometric Decision-based Attack (GeoDA)~\cite{DBLP:conf/cvpr/RahmatiMFD20}. We apply the same settings of attacks as Zhang~\etal~\cite{DBLP:journals/corr/abs-2111-15160}. To ensure controlled perturbations, we also constrain the magnitude of perturbation to a maximum of 1.0 in L2 norm. We show the results in Table~\ref{tab:result}.

\input{figures/result.tex}

{\em Initial success rate } represents the success rate of attacking the adversary's own copy. We apply the attacks on all the samples which are correctly classified by the adversary's copy. Then we compute of the percentage of adversarial samples which are successfully misclassified. Here we compute the percentage of successful adversarial attack for the model $\mathcal{M}_{\phi_1}$. {\em Replication rate on adversarial samples} measures the percentage of adversarial samples which can replicate on the other copy. Here we test two settings. Firstly, we test using the rewriter alone. This is done by applying adversarial samples generated on model $\mathcal{M}_{\phi_1}$ on model $\mathcal{M}_{\phi_2}$ since they share the same original model but rewritten using different attractors. Secondly, we test an approach which combines parameter rewriting and independent-training. This done by applying adversarial samples generated on model $\mathcal{M}_{\phi_1}$ on model $\mathcal{M}_{\psi_2}$ as they are rewritten using different attractors and their original models are also different.

From Table~\ref{tab:result}, we can observe that the attractor-based rewriter with flexible weight can significantly reduce the initial attack success rate for all three attacks. The replication rate of the proposed method is also lower. This observation is inline with the intuitions we discussed in Section~\ref{sec:improve}.

\subsection{Performance under Collusion Attack}
In this section, we evaluate the performance of the proposed method under collusion attacks. Firstly, we consider the custom attack in Zhang~\etal's work~\cite{DBLP:journals/corr/abs-2111-15160}. This attack is based on DeepFool~\cite{MoosaviDezfooli2016DeepFoolAS} and assumes a strong adversary who is able to reverse-engineer the model and obtain a white-box approximate. During an iteration of the attack, the adversary aggregates the outputs of all copies in collusion to decide the direction of movement. 
When the original attractor-based rewriter is applied, the attack success rate increases linearly with the number of copies used in collusion. Though the increase is gradual, the attack success rate will eventually reach 100\% when sufficiently large amount of copies are used.
\input{figures/newexp.tex}
The CIFAR-10 dataset was employed to evaluate the effectiveness of the aforementioned attack on the proposed approach. The results, depicted in Figure~\ref{fig:newcurve}, clearly demonstrate the superior performance of the new approach in the face of collusion attacks. Notably, the attack success rate does not exhibit unbounded growth but rather converges to approximately 15\%, even when forty copies are employed for collusion. This observation signifies the improved resilience of the new approach against collusion attacks compared to previous methods.

Secondly, we make modifications to boundary attack, so it can be evaluated under the setting of collusion.

Boundary attack selects an initial point that is distantly located and already classified as adversarial. It subsequently undertakes a random walk along the decision boundary region. This process ensures the sample remains adversarial while simultaneously minimizing the distance towards the original input image. In our modification to the algorithm, we require the sample to maintain its adversarial status across all colluding copies within each iteration.
We conduct the modified boundary attack on both the original attractor-based rewriter and the new rewriter with flexible weight, and present the outcomes in Figure~\ref{fig:boundary}.

\input{figures/boundary.tex}

From Figure~\ref{fig:boundary}, we can observe that the copies rewritten using the proposed method is more resistant against the collusion version of boundary attack. While the original attractor-based rewriter results a linear increase in attack success rate with respect to the number of colluding copies, the new method exhibits a much more gradual increase. Unfortunately, we found that the boundary attack is not capable of accommodating a large number of colluding parties. Therefore, we conduct experiments involving up to 10 colluding copies for the CIFAR-10 dataset and up to 8 colluding copies for the GTSRB dataset. At this point, for this attack, it remains uncertain whether the attack success rate for the new method is converging or slowly increasing.

%% file: figures/result.tex
\begin{table}[H]
    \vspace{-15pt}
    \small
    \centering
    \begin{tabular}{|c|c|c|c|c|c|c|c|} 
        \hline
                                                                                              & Datasets                                                                & \multicolumn{3}{c|}{CIFAR-10}                                              & \multicolumn{3}{c|}{GTSRB}                                                  \\ 
        \hline
        Approaches                                                                            & Attacks                                                                 & \begin{tabular}[c]{@{}c@{}}Boundary\\Attack\end{tabular} & HSJ    & GeoDA  & \begin{tabular}[c]{@{}c@{}}Boundary\\Attack\end{tabular} & HSJ    & GeoDA   \\ 
        \hline
        \multirow{2}{*}{\begin{tabular}[c]{@{}c@{}}Attractor\\(Fixed Weight)\end{tabular}}    & \begin{tabular}[c]{@{}c@{}}Initial\\Success\\Rate\end{tabular} & 59.1\%                                                   & 97.5\% & 94.8\% & 79.5\%                                                   & 96.4\% & 97.5\%  \\ 
        \cline{2-8}
                                                                                              & \begin{tabular}[c]{@{}c@{}}Replication\\Rate\end{tabular}               & 3.7\%                                                    & 2.5\%  & 9.2\%  & 0.7\%                                                    & 0.5\%  & 8.4\%   \\ 
        \hline
        \begin{tabular}[c]{@{}c@{}}Attractor\\ + Training \\(Fixed Weight)\end{tabular}       & \begin{tabular}[c]{@{}c@{}}Replication\\Rate\end{tabular}               & 3.7\%                                                    & 2.5\%  & 5.3\%  & 0.7\%                                                    & 0.1\%  & 2.3\%   \\ 
        \hhline{|========|}
        \multirow{2}{*}{\begin{tabular}[c]{@{}c@{}}Attractor\\(Flexible Weight)\end{tabular}} & \begin{tabular}[c]{@{}c@{}}Initial\\Success\\Rate\end{tabular} & 28.8\%                                                   & 44.3\% & 25.6\% & 44.3\%                                                   & 40.2\% & 32.4\%  \\ 
        \cline{2-8}
                                                                                              & \begin{tabular}[c]{@{}c@{}}Replication\\Rate\end{tabular}               & 3.2\%                                                    & 2.5\%  & 4.3\%  & 0.4\%                                                    & 0.2\%  & 2.0\%   \\ 
        \hline
        \begin{tabular}[c]{@{}c@{}}Attractor\\ + Training \\(Flexible Weight)\end{tabular}    & \begin{tabular}[c]{@{}c@{}}Replication\\Rate\end{tabular}               & 3.1\%                                                    & 2.5\%  & 4.2\%  & 0.4\%                                                    & 0.2\%  & 2.0\%   \\
        \hline
        \end{tabular}
    \caption{Performance under replication attacks.}
    \label{tab:result}
    \end{table}
    \vspace{-20pt}

%% file: figures/newexp.tex
\begin{figure}[H]
  \vspace{-15pt}
  \centering
      \centering
      \includegraphics[width=0.6\linewidth]{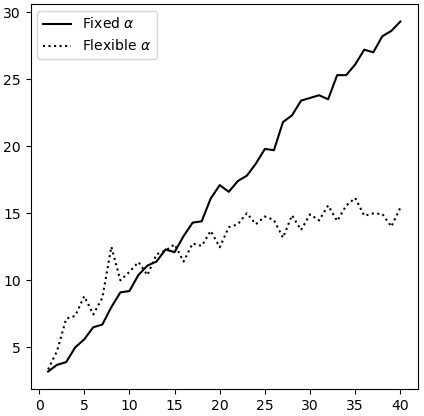}
    \caption{Performance under custom collusion attack.}
    \label{fig:newcurve}
  \end{figure}
  \vspace{-15pt}

%% file: figures/boundary.tex
\begin{figure}[H]
  \vspace{-15pt}
    \centering
    \begin{subfigure}{.5\linewidth}
      \centering
      \includegraphics[width=0.8\linewidth]{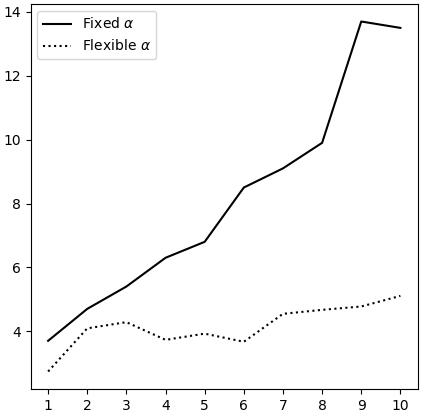}
      \caption{CIFAR-10.}
    \end{subfigure}%
    \begin{subfigure}{.5\linewidth}
      \centering
      \includegraphics[width=0.8\linewidth]{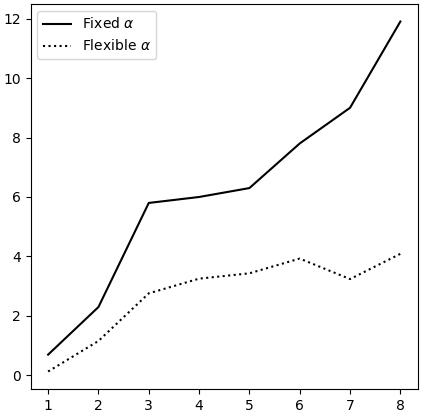}
      \caption{GTSRB.}
    \end{subfigure}
    \caption{Performance under boundary collusion attack.}
    \label{fig:boundary}
\end{figure}
\vspace{-15pt}

%% file: conclusion.tex
\section{Conclusion}

In conclusion, this paper addressed the collusion attack in the sell-buyer distribution setting of machine learning models. Through the use of parameter rewriting with an attractor-based rewriter, the existing state-of-the-art approach successfully distributed copies with different attractors, resulting in non-replicable adversarial samples across copies. However, in scenarios where multiple buyers collude to launch attacks, the collusion attack proved much stronger than independent attacks. To address this, we proposed the utilization of adaptive attractors guided by a U-shape curve to overcome the limitations of the existing method. Our findings highlight the potential of attractor-based defenses and the importance of adaptively injecting attractors according to each input sample. Future research can explore further advancements in adaptive attractors, for example, using a machine learning algorithm to control the injection of each individual hole and bump.